\pdfoutput=1
\documentclass[11pt]{article}
\usepackage[margin=1in]{geometry}
\usepackage[T1]{fontenc}
\usepackage[utf8]{inputenc}
\usepackage{lmodern}
\usepackage{microtype}
\usepackage{amsmath,amssymb}
\usepackage{booktabs}
\usepackage{makecell}
\usepackage{multirow}
\usepackage{graphicx}
\graphicspath{{figures/}}
\usepackage{adjustbox}
\usepackage{url}
\usepackage{hyperref}
\usepackage{placeins}
\usepackage[numbers,sort&compress]{natbib}
\hypersetup{colorlinks=false,pdfborder={0 0 0}}

\newcommand{\ket}[1]{\lvert #1 \rangle}
\newcommand{\bra}[1]{\langle #1 \rvert}

\title{\textbf{Algebraic Quantum Intelligence:\\A New Framework for Reproducible Machine Creativity}}
\author{%
  \parbox{0.95\textwidth}{\centering
    \small
    Kazuo Yano\textsuperscript{1,2*}, Jonghyeok Lee\textsuperscript{1}, Tae Ishitomi\textsuperscript{1}, Hironobu Kawaguchi\textsuperscript{1}, Akira Koyama\textsuperscript{1}, Masakuni Ota\textsuperscript{1}, Yuki Ota\textsuperscript{1}, Nobuo Sato\textsuperscript{1}, Keita Shimada\textsuperscript{1}, Sho Takematsu\textsuperscript{1}, Ayaka Tobinai\textsuperscript{1}, Satomi Tsuji\textsuperscript{1}, Kazunori Yanagi\textsuperscript{1}, Keiko Yano\textsuperscript{1}, Manabu Harada\textsuperscript{1}, Yuki Matsuda\textsuperscript{1}, Kazunori Matsumoto\textsuperscript{1}, Kenichi Matsumura\textsuperscript{1}, Hamae Matsuo\textsuperscript{1}, Yumi Miyazaki\textsuperscript{1}, Kotaro Murai\textsuperscript{1}, Tatsuya Ohshita\textsuperscript{1}, Marie Seki\textsuperscript{1}, Shun Tanoue\textsuperscript{1}, Tatsuki Terakado\textsuperscript{1}, Yuko Ichimaru\textsuperscript{1}, Mirei Saito\textsuperscript{1}, Akihiro Otsuka\textsuperscript{1}, Koji Ara\textsuperscript{1}\\[0.5ex]
    \normalsize
    \textsuperscript{1} Happiness Planet, Ltd., \textsuperscript{2} Hitachi, Ltd.\\
    \textsuperscript{*} kazuo.yano.bb@hitachi.com
  }%
}
\date{}

\begin{document}
\maketitle

\begin{abstract}
Large language models (LLMs) have achieved remarkable success in generating fluent and contextually appropriate text; however, their capacity to produce genuinely creative outputs remains limited. This paper posits that this limitation arises from a structural property of contemporary LLMs: when provided with rich context, the space of future generations becomes strongly constrained, and the generation process is effectively governed by near-deterministic dynamics. Recent approaches such as test-time scaling and context adaptation improve performance but do not fundamentally alter this constraint. To address this issue, we propose Algebraic Quantum Intelligence (AQI) as a computational framework that enables systematic expansion of semantic space. AQI is formulated as a noncommutative algebraic structure inspired by quantum theory, allowing properties such as order dependence, interference, and uncertainty to be implemented in a controlled and designable manner. Semantic states are represented as vectors in a Hilbert space, and their evolution is governed by C-values computed from noncommutative operators, thereby ensuring the coexistence and expansion of multiple future semantic possibilities. In this study, we implement AQI by extending a transformer-based LLM with more than 600 specialized operators. We evaluate the resulting system on creative reasoning benchmarks spanning ten domains under an LLM-as-a-judge protocol. The results show that AQI consistently outperforms strong baseline models, yielding statistically significant improvements and reduced cross-domain variance. These findings demonstrate that noncommutative algebraic dynamics can serve as a practical and reproducible foundation for machine creativity. Notably, this architecture has already been deployed in real-world enterprise environments.

\end{abstract}

\section{Introduction}

Creativity has traditionally been regarded as an intellectual activity dependent on chance, intuition, or individual genius, and has therefore not typically been treated as a phenomenon possessing reproducibility and controllability. This raises a fundamental question: Can creativity be mathematically designed? The present study addresses this question by proposing a theoretical framework and offering one possible answer regarding its applicability.

In the fields of artificial intelligence and machine learning, it has generally been assumed that learning from large amounts of data improves predictive accuracy, and such improvement has been considered desirable. This assumption indeed holds for problems in which correct answers exist in the past data and can be imitated. However, for problems in which no such imitable past solutions exist---referred to in this study as creative problems---this premise does not necessarily apply. Many critical decisions, including business management and strategic decision-making, policy formulation by governments and municipalities, and individual career planning, fall into this category of creative problems, where past optimal solutions cannot simply be reused.

As the use of generative AI continues to expand, the demand for AI-assisted support in such creative problems has increased rapidly. In recent years, large language models (LLMs) have achieved dramatic progress in their ability to integrate broad knowledge and generate fluent, contextually appropriate natural language~\cite{ref1,ref2,ref3,ref4}. At the same time, it has been pointed out that existing LLMs remain constrained in creative reasoning---understood as the ability to generate genuinely new conceptual structures or problem formulations that go beyond extensions of known knowledge and patterns~\cite{ref5}.

To address this limitation, numerous methods have been proposed, including test-time scaling to increase computational resources during inference~\cite{ref6,ref7,ref8}, context adaptation and in-context learning~\cite{ref9,ref10,ref11,ref12}, and self-reflective or iterative reasoning strategies~\cite{ref13}. While these approaches improve predictive accuracy and robustness, they do not fundamentally resolve the structural tendency of the generation process itself to fall into a closed and limited search space.

We argue that this limitation arises from the foundational assumptions underlying semantic generation in current LLMs. Specifically, LLMs are conditional generative models based on high-dimensional probability distributions, and when sufficiently long contexts are provided, future generations become highly constrained and nearly deterministic. While this property is advantageous for problems emphasizing predictability and reproducibility, it structurally collapses the space of semantic possibilities and makes it difficult to autonomously expand the scope of exploration. In other words, semantic generation in current LLMs is effectively governed by near-deterministic dynamics, which we contend inhibits the emergence of creativity.

In contrast, historical perspectives on human linguistic creativity and cultural evolution suggest that the evolution of meaning and concepts is not a continuous process uniquely determined by past states. Rather, it is better understood as a process in which multiple latent possibilities are simultaneously maintained and then diverge and expand through contextual interactions~\cite{ref14,ref15}. Research in cognitive science and social studies has shown that creative outcomes do not arise from simple combinations of individual elements, but instead emerge from nonlinear dynamics that depend on interaction order and contexts~\cite{ref16,ref17,ref18,ref19}. Such characteristics are difficult to capture within the frameworks of classical probability theory or static representational spaces.

In response to this challenge, recent studies in quantum probability theory and quantum cognitive models have demonstrated that contextuality, order effects, and interference phenomena in human judgment and concept combination can be effectively explained using quantum-theoretic formalismss~\cite{ref20,ref21,ref22}. In these models, semantic states are represented as state vectors in a Hilbert space, and systematic deviations from classical probability---arising from vector projections---give rise to context-dependent judgments and order effectss~\cite{ref23}. Similar non-classical probabilistic structures have been observed not only in human narrative texts but also in texts generated by large language modelss~\cite{ref24}, suggesting a structural correspondence between semantic processing and quantum-theoretic frameworks.

However, most existing quantum-inspired semantic and cognitive models focus on describing meaning composition and decision processes within fixed representational spaces, and do not provide design principles for dynamically expanding or reorganizing the semantic space itself~\cite{ref25,ref26,ref27,ref28,ref29,ref30}. As a result, no computational framework has yet been proposed that actively controls and exploits quantum-theoretic contextuality and interference as sources of creativity.

To bridge this gap, we propose Algebraic Quantum Intelligence (AQI). AQI introduces noncommutative algebraic structures from quantum theory as a central principle of semantic generation, thereby enabling systematic expansion of semantic space. AQI can be positioned as a type of Algebraic Quantum System (AQS), which generalizes Physical Quantum Systems (PQS) through noncommutative algebra, without assuming constraints tied to physical implementation such as decoherence or measurement limitations. This allows quantum-theoretic properties---such as order dependence, interference, and uncertainty---to be implemented in a fully designable and controllable manner.

In AQI, semantic states are represented as vectors in a Hilbert space, and their temporal evolution is described by the action of multiple noncommuting generative operators. Central to this framework is the idea of treating noncommutativity itself as a fundamental quantity of creativity. We refer to this measure as the Creativity Value (C-value). The magnitude of the discrepancy that arises when different perspectives or operations are applied in different orders directly corresponds to creative potential, understood as the breadth of future semantic branching. The basic relation \(C=\lvert\langle AB-BA\rangle\rvert\) provides a foundational equation linking creativity to conceptual operations and defines a structure that prevents semantic generation from collapsing onto a single deterministic trajectory.

In this study, we implement AQI as a system that realizes semantic generation based on noncommutative operator dynamics. The implementation introduces more than 600 specialized operators and constructs a dynamic semantic field in which operators are activated, suppressed, and reconfigured in response to contextual progression. We evaluate AQI on creative task benchmarks spanning ten domains using an LLM-as-a-judge protocol. The results show that AQI consistently achieves statistically significant performance improvements over strong existing baselines. Specifically, an average improvement of 27 points is observed in the creativity score (Co-Creativity Index; CCI), and the contribution of operator order dependence and interference effects to creative performance is empirically confirmed.

In summary, this work formalizes the limitations of creativity in current LLMs in terms of deterministic constraints on semantic generation, and introduces quantum-theoretic principles based on noncommutative algebra to enable dynamic expansion of semantic space. The proposed framework provides a new theoretical and computational foundation for treating machine creativity as a reproducible and designable phenomenon.

Finally, the architecture proposed in this study has been implemented not only as a theoretical construct but also as a commercial system deployed in real-world enterprise environments. Within a few months of launch, the system has been adopted by several dozen organizations, providing additional evidence that noncommutative semantic dynamics can deliver practical value in contexts requiring high-level decision-making. Operational and commercial details are omitted for reasons of confidentiality.

\section{Theoretical Framework and Implementation of Algebraic Quantum Intelligence (AQI)}

\subsection{Significance of Noncommutative Algebra}

The motivation for focusing on quantum-theoretic formulations---particularly quantum algebra and noncommutativity---stems from the hypothesis that creativity involves more than the mere accumulation of information or linear transmission of knowledge. Rather, it inherently entails discontinuous branching and context-dependent semantic transformations. In contrast, conventional AI systems, especially large language models (LLMs), while excelling in predictive accuracy and internal consistency, tend to exhibit convergent outputs as contextual information accumulates~\cite{ref1,ref2}. This convergence has been shown to constrain the generation of genuinely novel meanings or values---that is, creativity itself~\cite{ref31}.

This problem awareness closely resonates with research on creativity in human societies. In Csikszentmihalyi's systems model of creativity~\cite{ref16}, creative outcomes emerge from multilayered interactions among individuals, knowledge domains, and evaluative communities, where differences in order and context are essential for semantic expansion. Similarly, Nonaka and colleagues' SECI model~\cite{ref17} characterizes knowledge creation as a nonlinear process arising from cyclical and dynamic interactions rather than static accumulation. In addition, quantum cognitive models have demonstrated that context dependence and order effects in human decision-making and concept combination can be explained using Hilbert spaces and noncommutative structures~\cite{ref20,ref21,ref22,ref23}. Building on these insights, this study advances the hypothesis that quantum algebra and noncommutativity may provide a generative space that enables AI systems to transcend deterministic limitations and support creativity.

This chapter first clarifies the nature of deterministic convergence in LLMs and the essence of their creativity constraints. It then discusses how a quantum-theoretic framework based on noncommutative algebra yields what we term creativity with a guaranteed lower bound on branching width. We further articulate the bridge from theory to implementation, hierarchical dynamics governed by Hamiltonians, and the general framework of Algebraic Quantum Systems (AQS), which abstract physical quantum systems into a computationally designable form.

Modern large language models exhibit a structural limitation whereby richer contextual information leads to increasingly deterministic convergence of outputs, thereby intrinsically constraining creativity. This limitation arises from the generative principle of LLMs, which is grounded in sequential maximum-likelihood selection based on conditional probability distributions,
\begin{equation*}
P(x_{t+1}\mid x_1,\ldots,x_t),
\end{equation*} where \(x_{t}\) denotes the token generated at time \(t\)~\cite{ref1,ref2}. As the context length increases, the distribution P becomes sharply peaked, rendering the selection of the next token \(x_{t+1}\) effectively unique. Such deterministic convergence does not guarantee a fundamental width of semantic branching (e.g., creativity), even when random sampling or noise injection is applied. Consequently, LLMs are inherently prone to progressive reduction of their exploration space.

In AQI, semantic states are represented as state vectors in a Hilbert space 
\begin{equation*}
\ket{\psi}\in H.
\end{equation*}
This abstraction expresses the property of creative reasoning whereby meaning does not collapse to a single solution, but instead retains multiple latent possibilities simultaneously, represented as superposition.

The algebraic quantum intelligence (AQI) framework proposed in this study introduces noncommutative algebra to guarantee the emergence of creativity. Noncommutative algebra is defined by the property that, for operators A and B,
\begin{equation*}
AB\neq BA.
\end{equation*}
This implies that the order of application fundamentally alters the resulting semantic state. Within this framework, the state vector \(\ket{\psi}\) in Hilbert space H represents meaning, while semantic transformations are described by sequential applications of operators A,B,..., each corresponding to a specialized or characteristic perspective. Noncommutativity ensures that the choice of which operations are applied, and in what order, diversifies the generated states, thereby guaranteeing a nonzero branching width---that is, creativity---at a principled level.

To illustrate how noncommutativity directly gives rise to creativity, consider a concrete example involving two specialized operators: a financial perspective (A: Super CFO) and a human-resources perspective (B: Super CHRO). For a given new business initiative, applying the financial operator A first to assess financial soundness, followed by the human-resources operator B to optimize organizational structure (the sequence \(A\to B\)), yields a particular organizational design proposal. Conversely, applying B first to implement bold organizational reforms and then applying A to reallocate budgets (the sequence \(B\to A\)) leads to fundamentally different ideas and risk preferences, resulting in a distinct business strategy. If the AI model were commutative (AB=BA), changing the order would not affect the outcome, and the system would invariably converge to an ``average'' or ``safe'' plan. Strong noncommutativity, however, allows the same combination of expertise to produce entirely different creative branches depending on order. Indeed, our experiments observed markedly different output distributions between the \(A\to B\) and \(B\to A\) sequences (see Chapter 3). Thus, order dependence---noncommutativity---constitutes a fundamental principle by which AI systems can generate nontrivial creative leaps and structural transformations.

To quantify creativity, we define the Creativity Value (C-value) as the absolute value of the expectation of the commutator between operators,
\begin{equation*}
[A,B]=AB-BA.
\end{equation*}
The C-value is defined as 
\begin{equation*}
C=\left|\bra{\psi}\,[A,B] \,\ket{\psi}\right|\quad (\text{abbreviated as } C=\left|\langle AB-BA\rangle\right|).
\end{equation*}
The C-value is interpreted as the branching width of creativity for a given semantic state \(\ket{\psi}\), reflecting the divergence that arises when different capabilities A and B are applied in different orders. Larger C-values indicate greater potential for novel meaning generation through combinations of expertise, corresponding to greater creative potential.

Importantly, a larger branching width does not automatically imply higher-quality creativity. Expanding width alone may, in principle, generate meaningless diversity or noise. In AQI, noncommutativity (as measured by the C-value) guarantees the possibility of creative branching, but whether such branching yields qualitative transformation or valuable redefinition depends on the quality and composition of operators, as well as the historical and contextual dependencies of the semantic state. In this study, we evaluate whether increases in branching width are empirically associated with greater semantic depth and perspective expansion for recipients. Chapter 3 defines the evaluation metrics and experimentally demonstrates that such increases lead to higher evaluation scores. AQI is therefore designed to pursue reproducible creativity along both dimensions of width and quality.

Between such noncommuting operators, the following Robertson-type uncertainty relation holds~\cite{ref32}:
\begin{equation*}
\sigma_A\sigma_B \ge \frac{\left|\langle [A,B] \rangle\right|}{2},
\end{equation*}
where \(\sigma_A\) and \(\sigma_B\) denote the standard deviations of operators A and B, respectively. Here, in the context of semantic space, the standard deviation measures the extent to which the influence of an operator spreads across the state space. This relation thus provides a principled lower bound on creativity, implying that the simultaneous application of multiple forms of expertise necessarily induces dispersion in meaning. In other words, it guarantees a minimum level of creative divergence by construction (Fig. 1).

\begin{figure}[!tbp]
  \centering
  \includegraphics[
      width=0.95\linewidth,
      trim=0.6cm 5cm 0.2cm 4.5cm,
      clip
  ]{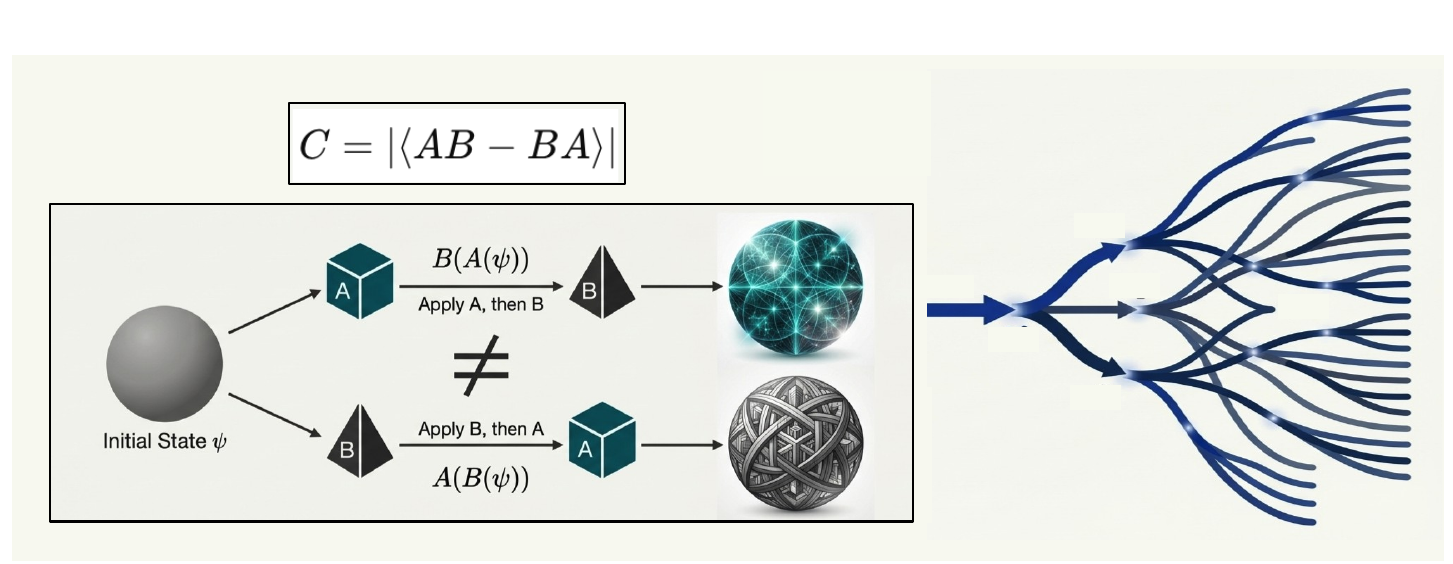}
  \caption{Conceptual semantic branching of non-commutative semantic dynamics in Algebraic Quantum Intelligence (AQI). While conventional LLMs exhibit convergence toward a single semantic trajectory as contextual constraints increase, AQI maintains multiple future semantic possibilities through non-commutative operator interactions, preventing premature collapse of exploration.}
  \label{fig:1}
\end{figure}

\subsection{Algebraic Quantum Systems (AQS)}

In this study, we construct the underlying theoretical framework using an abstract formulation termed the Algebraic Quantum System (AQS), which subsumes and generalizes conventional Physical Quantum Systems (PQS) based on physical quantum computers (Fig. 2).
\begin{figure}[!tbp]
  \centering
  \includegraphics[
      width=0.9\linewidth,
      trim=1.2cm 3.5cm 0cm 2.5cm,
      clip
  ]{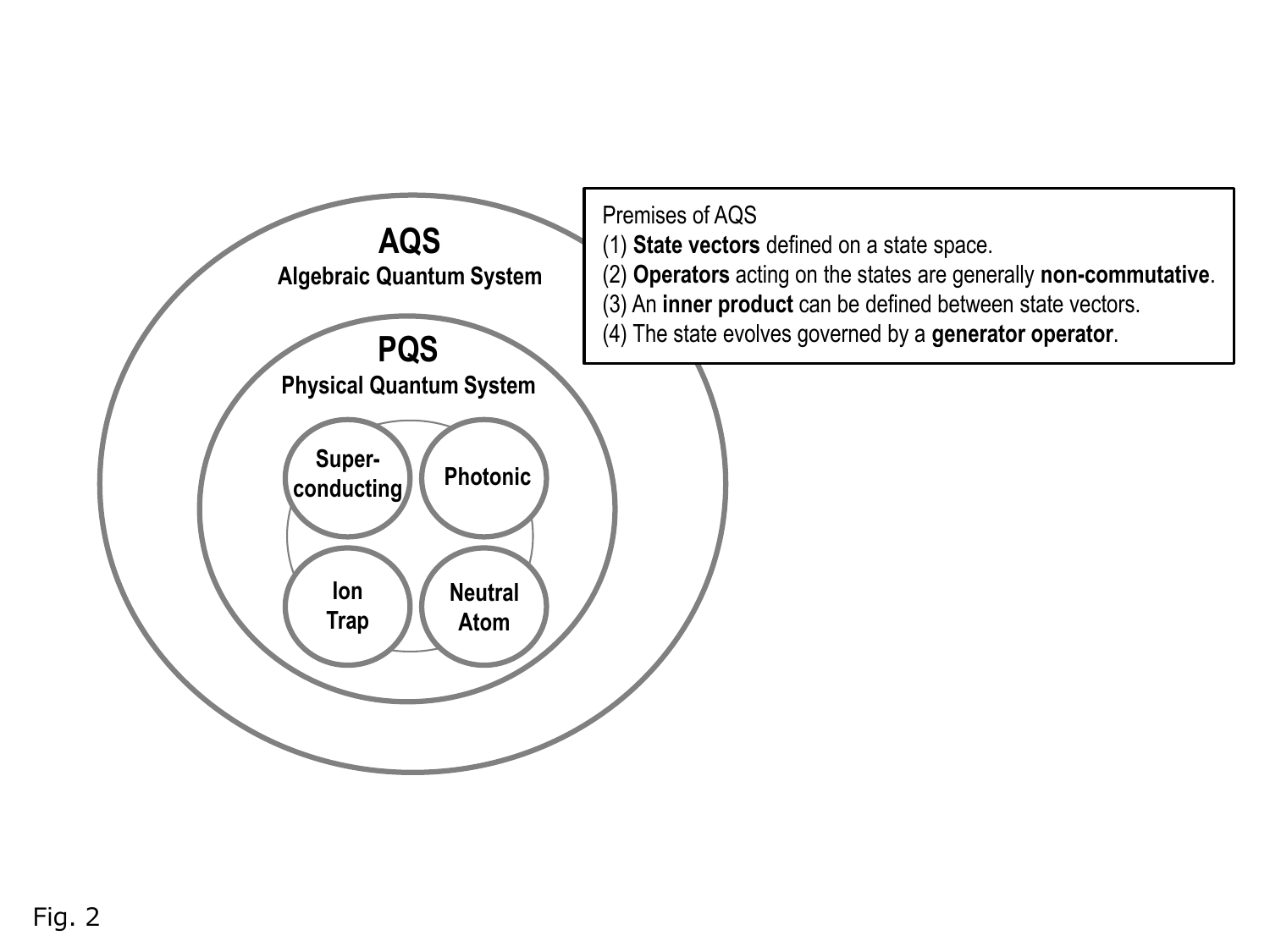}
  \caption{Algebraic Quantum System (AQS) as a generalization framework encompassing Physical Quantum Systems (PQS). AQS abstracts the minimal algebraic structures---state space, non-commutative operators, inner products, and generator-based dynamics---required to produce order dependence, interference, and uncertainty, without imposing physical constraints such as unitarity or measurement postulates.}
  \label{fig:2}
\end{figure}

PQS refers to information-processing systems implemented using qubits realized on various physical platforms, including superconducting circuits~\cite{ref33}, ion traps~\cite{ref34}, photonic systems~\cite{ref35}, and neutral atoms~\cite{ref36}, as well as quantum annealers~\cite{ref38,ref39}. In contrast, AQS generalizes these systems by extracting only the purely noncommutative algebraic structures---namely Hilbert spaces, operator algebras, and state evolution equations---independent of physical implementation. By doing so, AQS defines a universal quantum-theoretic system that is unconstrained by physical realizability and can be applied to information processing and AI creativity dynamics.

Here, we clarify the intended scope of the term quantum as used in this paper. AQI and its foundational Algebraic Quantum System (AQS) are not designed to faithfully simulate physical quantum systems. Rather, they adopt noncommutative operator algebras as design principles for semantic generation, abstracting away physical axioms such as measurement postulates, the Born rule, and unitary time evolution.

Even when focused on purely noncommutative algebraic structures independent of physical realization, significant degrees of freedom remain. In this work, we retain only the minimal assumptions required for noncommutativity and the emergence of uncertainty relations, while deliberately avoiding additional constraints. This choice is motivated by the hypothesis that uncertainty principles constitute a fundamental source of creativity.

An AQS satisfying these requirements is defined by the following assumptions:
\begin{enumerate}
\item It possesses state vectors \(\ket{\psi}\) (or, more generally, density matrices \(\rho\)) defined on a linear state space.
\item There exists a set of operators acting on this state space, which are generally noncommutative.
\item An inner product is defined between state vectors.
\item State evolution is governed by a generative operator H.
\end{enumerate}

State updates may take the form of a generalized quantum master equation such as
\begin{equation*}
\ket{\psi(t+\Delta)} = H(t)\,\ket{\psi(t)},
\end{equation*}
where \(\Delta\) denotes an update unit and H(t) is a time-dependent evolution operator. More generally, the framework encompasses state-dependent update mappings induced by H(t)~\cite{ref40,ref41,ref42}.

We initially considered adopting the Schr\"odinger equation---the cornerstone of quantum mechanics~\cite{ref43},
\begin{equation*}
i\,\frac{d}{dt}\ket{\psi} = H\,\ket{\psi},
\end{equation*}
as a foundational assumption. However, because the Schr\"odinger equation presupposes the existence of a conserved quantity (energy), it would impose overly restrictive constraints on the degrees of freedom of AQS. We therefore adopt the more generalized formulation above. This allows AQS to encompass open systems in which energy is not conserved, systems with discrete-step evolution, and systems for which defining energy as a conserved quantity is itself difficult. Such generality is essential for modeling the evolution of creative semantic states.

Although the state space is defined to have a linear structure, the time-dependent generator H(t) and the induced state updates are generally state-dependent and may involve nonlinear transformations in implementation.

Given the four assumptions above, the Robertson-type uncertainty principle~\cite{ref27} holds, and core quantum phenomena such as order dependence, interference, and superposition naturally emerge. In this sense, AQS extracts only the algebraic structures responsible for these quantum effects---namely, the existence of noncommutative operators and generator-driven state evolution.

Unlike physical quantum systems, however, AQS does not require assumptions such as unitary time evolution, measurement axioms, the Born rule, or physically observable quantities~\cite{ref44}. Conversely, physical quantum systems (PQS) can be understood as special cases of AQS obtained by imposing additional physical constraints such as unitarity and probabilistic interpretation.

This containment relationship can be schematically expressed as:
\begin{equation*}
\text{AQS (Algebraic Quantum System)}\supset\text{PQS (Physical Quantum System)}
\end{equation*}
\begin{equation*}
\supset\{\text{superconducting qubits},\ \text{photonic qubits},\ \text{neutral-atom qubits},\ \ldots\}.
\end{equation*}
That is, within the broad abstract framework of AQS lies PQS, and within PQS lie specific physical implementations such as superconducting, photonic, ionic, and neutral-atom quantum computers.

Within this hierarchy, AQI constitutes a specific instance of AQS, enabling the design and utilization of quantum-like features and noncommutative dynamics without being constrained by the physical realizability of quantum systems.
The dynamics of state evolution are described by a discretized and abstracted form of Schr\"odinger-type evolution governed by a Hamiltonian:
\begin{equation*}
\ket{\psi_{k+1}} = H(k)\,\ket{\psi_k}.
\end{equation*}
Here, the creative Hamiltonian H(k) has the structure
\begin{equation*}
H(k)=\sum_i \epsilon_i(k)\, a_i^{\dagger}a_i + \sum_{i,j} g_{ij}(k)\, a_i^{\dagger}a_j,
\end{equation*}
where \(a_i^{\dagger}\) and \(a_i\) are creation and annihilation operators corresponding to specialization \(i\)~\cite{ref45}. The suffix \(k\) denotes the progression of updates, and the coefficients \(\epsilon_i(k)\) and \(g_{ij}(k)\) are dynamically determined based on the current context, recent Creativity Values, and external factors. This Hamiltonian represents the integrated effect of the noncommutative set of specialized operators applied during the generation process.

Through this mechanism, AQI dynamically controls which specializations are applied, in what order, and with what intensity, thereby maximizing branching width (Creativity Value) while guaranteeing a structural lower bound on creativity.

\subsection{Implementation of AQI}

A central challenge was whether this mathematical algebraic system could be implemented as an AI system capable of handling language and meaning in practice.

In the concrete implementation, the abstract elements of AQI are instantiated as follows. The semantic state \(\ket{\psi}\) corresponds to the contextual representations maintained inside a language model, including its context vectors. Operators A, B correspond to specialized transformation modules or prompt-level operators that modify these representations. Expectation values are estimated via inner-product computations over controlled generations; cosine similarity between vectors or correlation coefficients (i.e., cosine similarity centered by the mean) can be used as practical inner products.

To realize AQI in an operational AI system, we adopt a two-layer architecture consisting of semantic state updates and operator updates (Fig. 3). Updates are performed at the granularity of a message, defined as a coherent semantic and contextual unit composed of one or more paragraphs, within which shifts in expertise, perspective, or branching may occur.

\begin{figure}[!tbp]
\centering
\includegraphics[
    width=0.8\linewidth,
    trim=0cm 6cm 0cm 5.5cm,
    clip
]{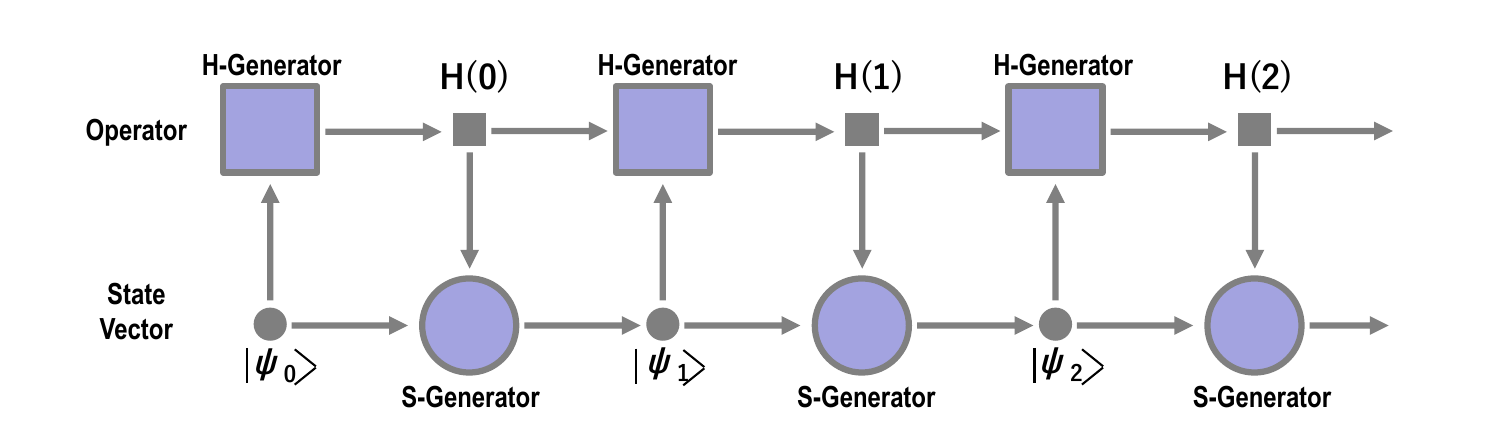}
\caption{Two-layer architecture of AQI: alternating state update and operator update. The semantic state \(\ket{\psi_k}\) is updated by a dynamically generated creative Hamiltonian H(k) (S-Generator), while the Hamiltonian itself is adaptively constructed based on the current state and prior dynamics (H-Generator), enabling non-commutative semantic evolution across message units.}
\label{fig:3}
\end{figure}

Let \(\ket{\psi_k}\) denote the state of the semantic space at step k. This state is updated to \(\ket{\psi_{k+1}}\) by the creative Hamiltonian H(k), which governs semantic evolution. In a deep-learning-based LLM system, the state vector \(\ket{\psi_k}\) is mapped to the entire context currently provided to and maintained within the model, together with the corresponding internal representations induced by that context.

To implement the update \(\ket{\psi_k}\to\ket{\psi_{k+1}}\) we developed a mechanism called the S-Generator (S for state). In addition, we developed a mechanism that generates and updates the creative Hamiltonian H(k), referred to as the H-Generator (H for Hamiltonian). The generative operator H(k) produced by the H-Generator functions, in an LLM-based system, as a mechanism that takes \(\ket{\psi_k}\) as input and generates the context required to produce \(\ket{\psi_{k+1}}\), while also referencing the previous Hamiltonian H(k-1).

In this way, the current semantic state \(\ket{\psi_k}\) is used to update the immediately preceding creative Hamiltonian H(k-1) into H(k). By iteratively alternating between the two layers---semantic state \(\ket{\psi_k}\) and operator H(k)---the system implements semantic evolution driven by Hamiltonian dynamics (Fig. 3). Both the S-Generator and the H-Generator operate through a coordinated interaction between procedural programs and LLM components.

Although the state vector \(\ket{\psi}\) in the quantum-theoretic framework is an element of a Hilbert space over the complex field \(\mathbb{C}\), practical deep-learning implementations represent state vectors in finite-dimensional spaces over the real field \(\mathbb{R}\). Core structural features required by the quantum framework---such as superposition, linear operations, and orthogonality---can be fully implemented in real-valued spaces \(\mathbb{R}^d\), and no essential barrier arises from this choice~\cite{ref40}. Moreover, complex Hilbert spaces can be reconstructed using pairs of real vectors or phase-augmented representations, posing no fundamental barrier to AI implementations.

Furthermore, conventional LLMs---that is, deterministic generative models---are contained as special cases within the AQI framework~\cite{ref24,ref25}. When all specialized operators commute, or when the operator sequence is fixed, state evolution becomes isomorphic to deterministic sequence generation governed by the conditional probability distributions of LLMs. AQI therefore subsumes the representational capacity of LLMs while extending it with a new dimension of noncommutativity and creativity.

The theoretical framework presented in this chapter provides an end-to-end connection---from noncommutative algebraic principles at the mathematical level, through deep-learning implementation and a two-layer hierarchical architecture, to the generalized AQS formulation---aimed at overcoming the fundamental constraint of exploration-space collapse in LLMs. This integrated architecture makes it practically feasible to move beyond purely data-driven knowledge expansion and toward AI systems endowed with the essential capability to design, control, and reproduce creativity. Subsequent chapters detail applications and evaluation metrics based on this theoretical foundation.

\section{Experiments and Results}

This chapter evaluates the proposed AQI framework with respect to three objectives:

(1) whether AQI yields measurable improvements in creative performance;
(2) whether such improvements arise from dynamics induced by noncommutative generators rather than from mere sampling randomness; and
(3) whether interference-like effects associated with noncommutativity are empirically observable.

Our central hypothesis is that the creativity limitations of current LLMs stem from the fact that semantic generation is effectively governed by near-deterministic behavior. AQI is distinguished by introducing noncommutative algebra, which renders the breadth of exploration explicitly designable through the C-value.

To test this hypothesis, we conduct the following three experiments:
\begin{enumerate}
\item[(E1)] Creativity Evaluation: Does AQI consistently outperform state-of-the-art baseline models across creative reasoning tasks spanning ten domains?
\item[(E2)] Effects of Noncommutativity in Quantum Algebra: Under identical conditions, does changing only the order of operator application produce systematically different output distributions?
\item[(E3)] Effects of Quantum Interference: Does the composition of sequential operations exhibit statistical properties that cannot be explained by simple (commutative) vector composition?
\end{enumerate}

\subsection{Creativity Evaluation}

The objective of AQI is not to reproduce previously learned knowledge, but to expand exploration within semantic space and generate responses with high creativity.

The notion of creativity considered here is not limited to narrow artistic domains such as art or music. Rather, we target creativity as required in real-world business and policy decision-making and complex problem solving. The ability to demonstrate creativity in such contexts carries significant social and economic impact, and the value of creativity in these settings is correspondingly high.

A fundamental difficulty in such tasks is that, although past information and knowledge may be referenced, there is no single correct answer embedded in the training data. Examples of such important open-ended tasks include business risk assessment, investment decision-making, financial strategy, corporate transformation, human resource strategy, sales strategy, technology strategy, partnership decisions, investor relations, and leadership practice.

Existing AI benchmarks do not adequately capture these real managerial challenges. For instance, MMLU, MATH, and GSM8K focus on academic problems with well-defined correct answers, while BIG-Bench, SuperGLUE, and BusinessBench address practical tasks but do not cover complex, high-impact corporate management problems.

To address this gap, we developed a new evaluation benchmark targeting socially valuable, creativity-intensive tasks. We designed ten domains of creative managerial reasoning that emphasize business value and require integrative thinking (Table 1). Each task is formulated as an open-ended question involving intertwined managerial, business, societal, cultural, and technological factors, informed by interviews with practicing executives. The ten domains are: 1. Risk and Strategic Foresight, 2. Investor Relations, 3. Inter-Organizational Collaboration, 4. Corporate Finance Optimization, 5. Investment Decision-Making, 6. Sales and Client Engagement, 7. Organizational Transformation, 8. Human Resource Strategy, 9. Technology Foresight and R\&D Direction, 10. Leadership and Mentorship. An overview of these tasks is provided in Table 1.

\begin{table}[t]
  \centering
  \caption{Creative reasoning benchmark domains and representative task descriptions.}
  \label{tab:aqi_table1_domains}
  \small
  \begin{adjustbox}{max width=\linewidth}
  \begin{tabular}{@{}ll@{}}
    \toprule
    Question Category & Subject \\
    \midrule
    1 Risk and Strategic Foresight & Regarding business risks over the next three years \\
    2 Investor Relations & Regarding explanations to overseas shareholders \\
    3 Inter-Organizational Collaboration & Regarding the selection of investment targets as a corporate venture capital (CVC) \\
    4 Corporate Finance Optimization & Regarding improvement of the cash conversion cycle \\
    5 Investment Decision-Making & Regarding the pros and cons of upfront investment in AI products \\
    6 Sales and Client Engagement & Regarding sales of systems for pharmaceutical companies \\
    7 Organizational Transformation & Regarding growth strategies for SaaS ventures \\
    8 Human Resource Strategy & Regarding labor shortages at regional companies \\
    9 Technology Foresight and R\&D Direction & Regarding the technology strategy of IT businesses for automotive companies \\
    10 Leadership and Mentorship & Regarding supporting the growth of a subordinate with gaps in their skillset \\
    \bottomrule
  \end{tabular}
  \end{adjustbox}
\end{table}

Responses to these tasks were evaluated along two axes. The first axis assesses the creativity of the AI-generated content itself, referred to as the \(C_x\) score. The \(C_x\) score is computed as the average of three criteria: novelty, surprise, and depth. These criteria reflect established components of creativity while preventing mere novelty from being overvalued.

The second axis evaluates whether the response enhances the creativity of the recipient, referred to as the \(C_y\) score. The \(C_y\) score assesses creativity expansion from the recipient's perspective, averaging four criteria: metacognitive stimulation, degree of reframing, autonomy enhancement, and personal engagement. This axis captures whether the response broadens perspectives, elevates thinking, and motivates action by making the problem personally salient.

High performance on both axes---indicating that both the AI and the human become more creative---defines the Co-Creativity Index (CCI). CCI is conceptualized as a two-dimensional plane spanned by \(C_x\) and \(C_y\), where higher values correspond to points closer to the upper-right corner (Fig. 4). To ensure that high performance on only one axis does not dominate the overall score, CCI is computed as \(\text{CCI} = \text{average}[\min(C_x, C_y)]\)~\cite{ref46}.

\begin{figure}[!tbp]
\centering
\includegraphics[
    width=0.8\linewidth,
    trim=2.5cm 2.8cm 2cm 2cm,
    clip
]{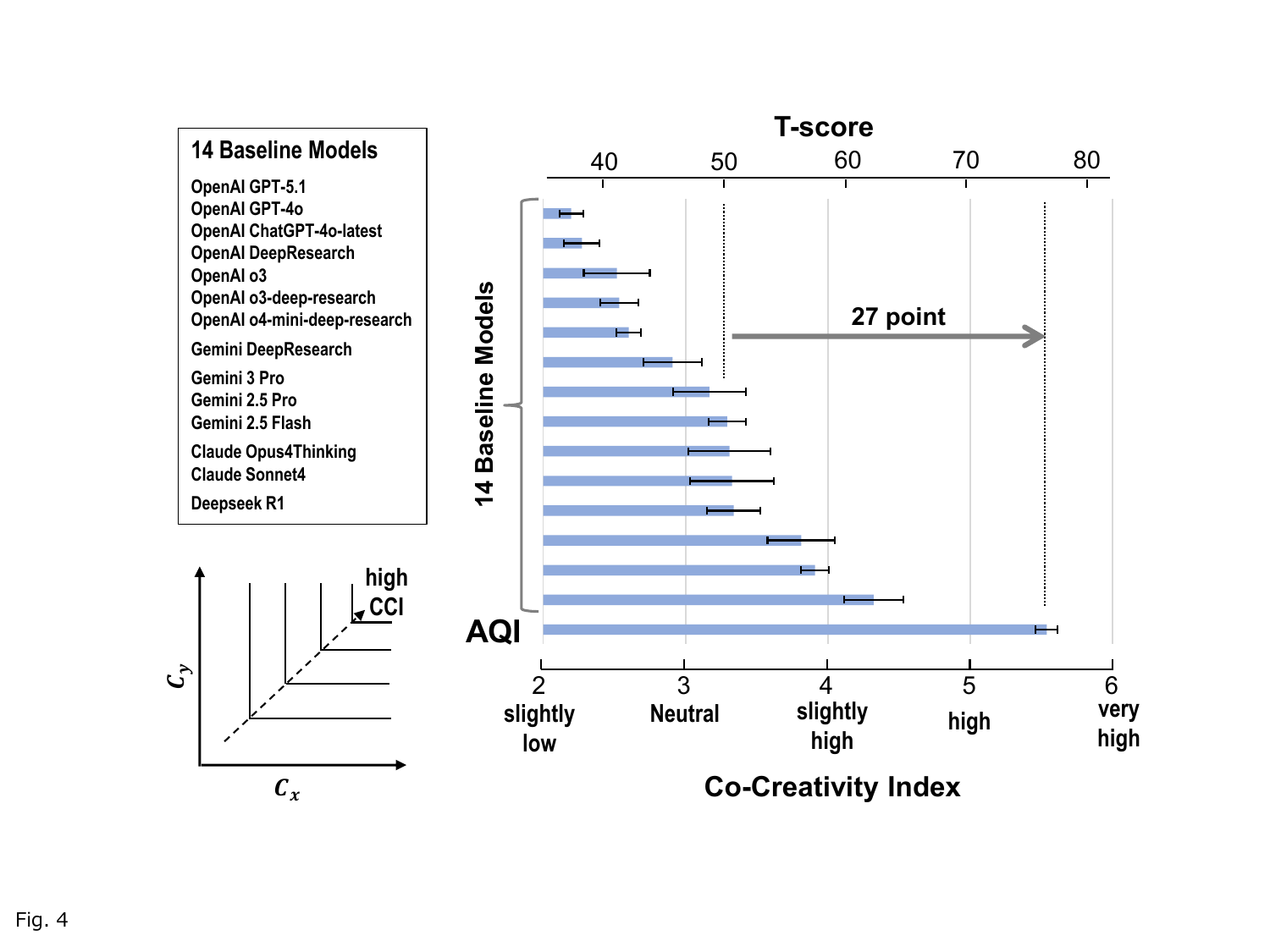}
\caption{Performance comparison across ten creative reasoning domains measured by the Co-Creativity Index (CCI). AQI consistently outperforms 14 strong baseline models, showing an average improvement of +27 T-score points while exhibiting reduced variance across domains, indicating both higher creativity and greater stability.}
\label{fig:4}
\end{figure}

Each dimension is rated on an eight-point Likert scale (1: very low -- 8: extremely high), and scores are aggregated to enable statistical comparison.

We adopted a standard LLM-as-a-judge evaluation protocol. Responses were scored under blind conditions by evaluator models (the average of GPT-4o and o3), with model identities concealed. All models were evaluated using identical prompt sets.

Fourteen models were included as baselines: GPT-5.1, GPT-4o, ChatGPT-4o-latest, DeepResearch, o3, o3-deep-research, o4-mini-deep-research, Gemini 3 Pro, Gemini 2.5 Pro, Gemini 2.5 Flash, Gemini DeepResearch, Claude 3 Opus 4 Thinking, Claude 3 Sonnet 4, DeepSeek R1

Across all ten domains, AQI outperformed the baseline models by an average of 27 T-score points on the CCI metric (Table 2, Fig. 4). Particularly notable improvements were observed in Sales and Client Engagement and Technology Foresight \& R\&D Direction, where AQI achieved T-scores exceeding 78. Moreover, AQI exhibited the smallest variance in scores across tasks, indicating stable performance across domains. These results suggest that AQI does not merely generate stylistically diverse responses, but consistently produces creative outputs characterized by problem reframing and integrative depth.

\begin{table*}[p]
  \centering
  \caption{Co-Creativity Index (CCI) results across ten domains for AQI and baseline models.}
  \label{tab:aqi_table2_cci}
  \setlength{\tabcolsep}{4pt}
  
  \vspace{0.5em}
  \begin{adjustbox}{max width=\textwidth}
  \begin{tabular}{@{}l*{6}{c}@{}}
    \toprule
    Models
      & \makecell{1\\Risk \&\\Strategic\\Foresight}
      & \makecell{2\\Investor\\Relations}
      & \makecell{3\\Inter-Organizational\\Collaboration}
      & \makecell{4\\Corporate Finance\\Optimization}
      & \makecell{5\\Investment\\Decision-Making} \\
    \midrule
    AI 1 & 2.17 & 2.13 & 2.00 & 2.13 & 2.17 \\
    AI 2 & 2.38 & 1.63 & 2.13 & 2.38 & 2.67 \\
    AI 3 & 2.83 & 2.00 & 2.00 & 1.88 & 2.75 \\
    AI 4 & 2.67 & 2.17 & 2.25 & 1.63 & 2.67 \\
    AI 5 & 2.38 & 2.25 & 3.00 & 3.00 & 2.38 \\
    AI 6 & 2.38 & 2.38 & 3.00 & 3.17 & 4.00 \\
    AI 7 & 3.38 & 2.13 & 2.63 & 3.00 & 4.50 \\
    AI 8 & 3.75 & 2.75 & 2.38 & 3.13 & 3.63 \\
    AI 9 & 3.00 & 4.50 & 1.63 & 2.38 & 4.50 \\
    AI 10 & 3.13 & 2.63 & 4.38 & 3.67 & 3.67 \\
    AI 11 & 3.83 & 2.63 & 3.13 & 2.38 & 4.17 \\
    AI 12 & 4.63 & 2.75 & 3.75 & 3.13 & 4.50 \\
    AI 13 & 4.38 & 3.50 & 4.00 & 3.63 & 3.63 \\
    AI 14 & 4.63 & 4.00 & 5.00 & 4.75 & 3.13 \\
    \midrule
    \textbf{AQI (This Work)} & \textbf{5.25} & \textbf{5.13} & \textbf{5.88} & \textbf{5.38} & \textbf{5.38} \\
    \midrule
    \addlinespace
    \midrule
    Average & 3.38 & 2.84 & 3.14 & 3.04 & 3.58 \\
    Standard Deviation & 0.95 & 0.97 & 1.19 & 0.99 & 0.91 \\
    T-score & 69.7 & 73.6 & 73.0 & 73.6 & 69.8 \\
    \midrule
  \end{tabular}
  \end{adjustbox}

  \vspace{1em}
  
  \begin{adjustbox}{max width=\textwidth}
  \begin{tabular}{@{}l*{7}{c}@{}}
    \toprule
    Models
      & \makecell{6\\Sales \& Client\\Engagement}
      & \makecell{7\\Organizational\\Transformation}
      & \makecell{8\\Human Resource\\Strategy}
      & \makecell{9\\Technology Foresight\\and R\&D Direction}
      & \makecell{10\\Leadership\\and Mentorship}
      & Average
      & \makecell{Standard\\Deviation} \\
    \midrule
    AI 1 & 2.13 & 2.67 & 2.67 & 1.75 & 2.17 & 2.20 & 0.26 \\
    AI 2 & 2.00 & 2.88 & 2.75 & 1.75 & 2.17 & 2.27 & 0.40 \\
    AI 3 & 2.50 & 2.25 & 2.63 & 2.00 & 4.50 & 2.53 & 0.73 \\
    AI 4 & 2.25 & 3.00 & 2.63 & 2.75 & 3.17 & 2.52 & 0.43 \\
    AI 5 & 2.25 & 2.50 & 2.67 & 2.75 & 2.83 & 2.60 & 0.28 \\
    AI 6 & 3.33 & 3.17 & 2.83 & 1.50 & 3.33 & 2.91 & 0.65 \\
    AI 7 & 2.88 & 4.13 & 4.63 & 2.63 & 3.50 & 3.34 & 0.80 \\
    AI 8 & 3.00 & 3.33 & 3.17 & 2.88 & 3.67 & 3.17 & 0.42 \\
    AI 9 & 2.75 & 4.33 & 3.25 & 3.00 & 3.75 & 3.31 & 0.91 \\
    AI 10 & 2.38 & 1.50 & 3.83 & 3.25 & 4.83 & 3.33 & 0.93 \\
    AI 11 & 3.38 & 3.33 & 3.17 & 2.75 & 4.17 & 3.29 & 0.59 \\
    AI 12 & 3.00 & 4.63 & 4.00 & 3.00 & 4.75 & 3.81 & 0.75 \\
    AI 13 & 3.75 & 3.83 & 4.00 & 3.88 & 4.50 & 3.91 & 0.31 \\
    AI 14 & 3.88 & 4.50 & 5.13 & 3.38 & 4.83 & 4.32 & 0.66 \\
    \midrule
    \textbf{AQI (This Work)} & \textbf{5.63} & \textbf{5.50} & \textbf{5.75} & \textbf{5.63} & \textbf{5.88} & \textbf{5.54} & \textbf{0.24} \\
    \midrule
    \addlinespace
    \midrule
    Average & 3.01 & 3.44 & 3.54 & 2.86 & 3.87 & 3.27 & 0.33 \\
    Standard Deviation & 0.90 & 1.02 & 0.96 & 0.98 & 1.01 & 0.85 & 0.08 \\
    T-score & 79.2 & 70.3 & 73.1 & 78.4 & 69.8 & 76.7 & 3.26 \\
    \midrule
  \end{tabular}
  \end{adjustbox}
\end{table*}

\subsection{Effects of Noncommutativity in Quantum Algebra}

We next examine the concrete effects of noncommutativity in AQI's quantum-algebraic formulation and distinguish them from mere increases in sampling randomness.

To this end, we compared conditions in which only the order of operator application differed, while all other factors were held constant. Specifically, we employed two specialized managerial operators:
\begin{quote}
A: ``Super CFO''\\
B: ``Super CHRO''
\end{quote}
and evaluated two sequences, \(A\to B\) and \(B\to A\), on the same task.

To suppress incidental variability, each condition was repeated 100 times under identical settings, with temperature fixed at 0 to minimize sampling-induced noise. Although some intrinsic nondeterminism remains within the model, this setup ensures that observed differences are not attributable to temperature-driven randomness.

For analysis, each output text was converted into a 3,072-dimensional embedding vector, followed by dimensionality reduction to two dimensions using principal component analysis. The resulting point distributions were then compared across the two order conditions (Fig. 5).

As shown in Fig. 5, the \(A\to B\) and \(B\to A\) conditions formed clearly separated, non-overlapping clusters. This demonstrates that even under identical tasks and conditions, changing only the order of operator application systematically generates distinct semantic trajectories---evidence that AQI's generators act in a noncommutative manner.

\begin{figure}[!tbp]
\centering
\includegraphics[
    width=0.7\linewidth,
    trim=2.2cm 1.8cm 1.5cm 0.5cm,
    clip
]{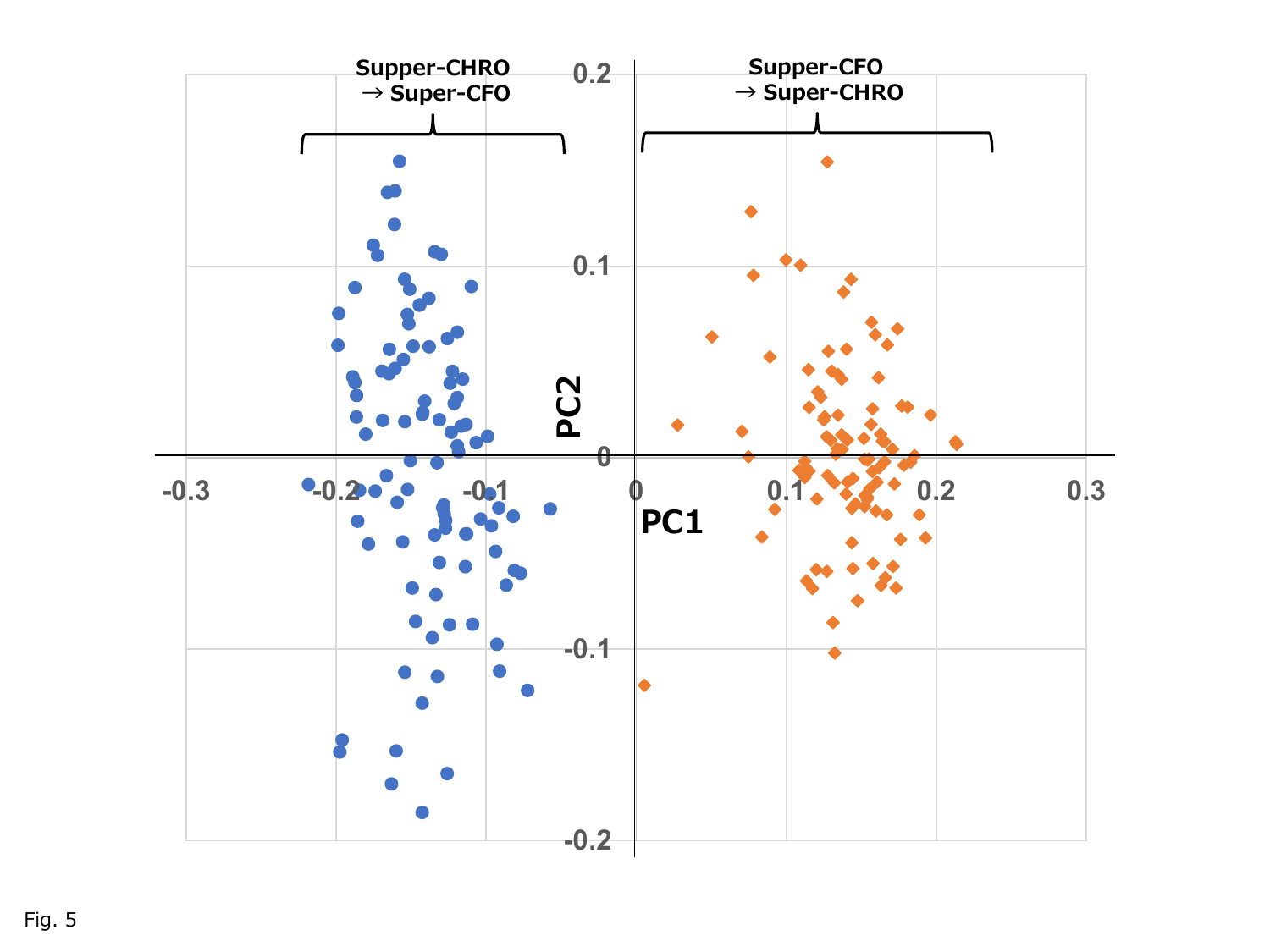}
\caption{Order-dependent semantic divergence induced by non-commutative operators. Embedding projections (PCA) of outputs generated under two conditions---\(A\to B\) (Super CFO \(\to\) Super CHRO) and \(B\to A\)---form clearly separated clusters even at temperature = 0, demonstrating systematic trajectory bifurcation driven solely by operator order.}
\label{fig:5}
\end{figure}

These findings align with the central hypothesis of AQI: even in regimes where LLM generation would otherwise converge deterministically, noncommutative generators induce trajectory branching. The separation observed at temperature = 0 further supports the conclusion that the performance improvements observed in (E1) is not a byproduct of added randomness, but a consequence of order-dependent dynamics.
From a theoretical perspective, this phenomenon can be interpreted as the manifestation of uncertainty in quantum algebra. Noncommutativity between operators yields a finite, nonzero C-value, which---via the Robertson-type inequality---imposes a lower bound on standard deviations. In the AQI context, this dispersion is interpreted not as noise, but as a guaranteed lower bound on exploration width in semantic space. Thus, the coexistence of noncommutative operators functions as a design principle that suppresses excessive collapse of exploration and ensures a minimum degree of creative branching.

\subsection{Effects of Quantum Interference}

Another major consequence of noncommutative operator action is quantum interference. Quantum interference refers to the emergence of rich, structured patterns when the effects of two operators are composed---patterns that cannot be explained by simple averaging or by commutative vector composition. We now examine how such effects manifest in the outputs of AQI.

For a given theme, let \(Y\) denote the response vector produced by the CFO operator alone, and let \(Y^{\prime}\) denote the response vector produced by the CFO operator after the CHRO operator has first been applied and its influence incorporated. Let \(X\) denote the response vector produced by the CHRO operator alone. Using the original CFO response \(Y\) as a reference, we define the following difference vectors:
\begin{equation*}
\Delta Y = Y^{\prime} - Y,
\end{equation*}
\begin{equation*}
\Delta X = X - Y.
\end{equation*}
Here, \(\Delta Y\) represents the change in the CFO response induced by the CHRO response X, while \(\Delta X\) represents how different the CHRO response is from the original CFO response, and thus serves as a vector encoding the ``cause'' of the change relative to the vector Y. Under a commutative and linear composition model, one would expect the change vector \(\Delta Y\) to correlate with the causal vector \(\Delta X\). Denoting the inner product between standardized vectors (i.e., the correlation coefficient) by \(\mathrm{corr}(\cdot)\), we define:
\begin{equation*}
r=\mathrm{corr}(Y^{\prime}, X),
\end{equation*}
\begin{equation*}
r^{\prime}=\mathrm{corr}(\Delta Y,\Delta X)=\mathrm{corr}(Y^{\prime}-Y,\ X-Y).
\end{equation*}
In general, one would expect \(r^{\prime} \ge r\), both from a causal perspective and because subtracting the same reference vector Y tends to increase shared components and thus correlation.

Empirically, however, we consistently observed \(r^{\prime}<r\) across all ten benchmark domains (Table 3). On average, while \(r \approx 0.85\), \(r^{\prime}\) dropped sharply to approximately 0.29--0.32. Such a decrease is difficult to explain using simple commutative composition models.

To verify that this discrepancy was not accidental, we performed a permutation test in which the components of the reference vector \(Y\) were randomly shuffled to produce a surrogate vector \(Y_r\). We then computed:
\begin{equation*}
r^{\prime\prime}=\mathrm{corr}(Y^{\prime}-Y_r,\ X-Y_r).
\end{equation*}
In this case, \(r^{\prime\prime}\) averaged approximately 0.90, exhibiting behavior consistent with commutative intuition (Table 3, Fig. 6). This result indicates that the empirically observed condition \(r^{\prime}<r\) in the real data is not due to statistical noise or computational artifacts, but instead reflects an interference-like structure in which component-wise amplification and suppression occur simultaneously.

\begin{figure}[!tbp]
\centering
\includegraphics[
    width=0.6\linewidth,
    trim=4.2cm 2.8cm 4.2cm 3.3cm,
    clip
]{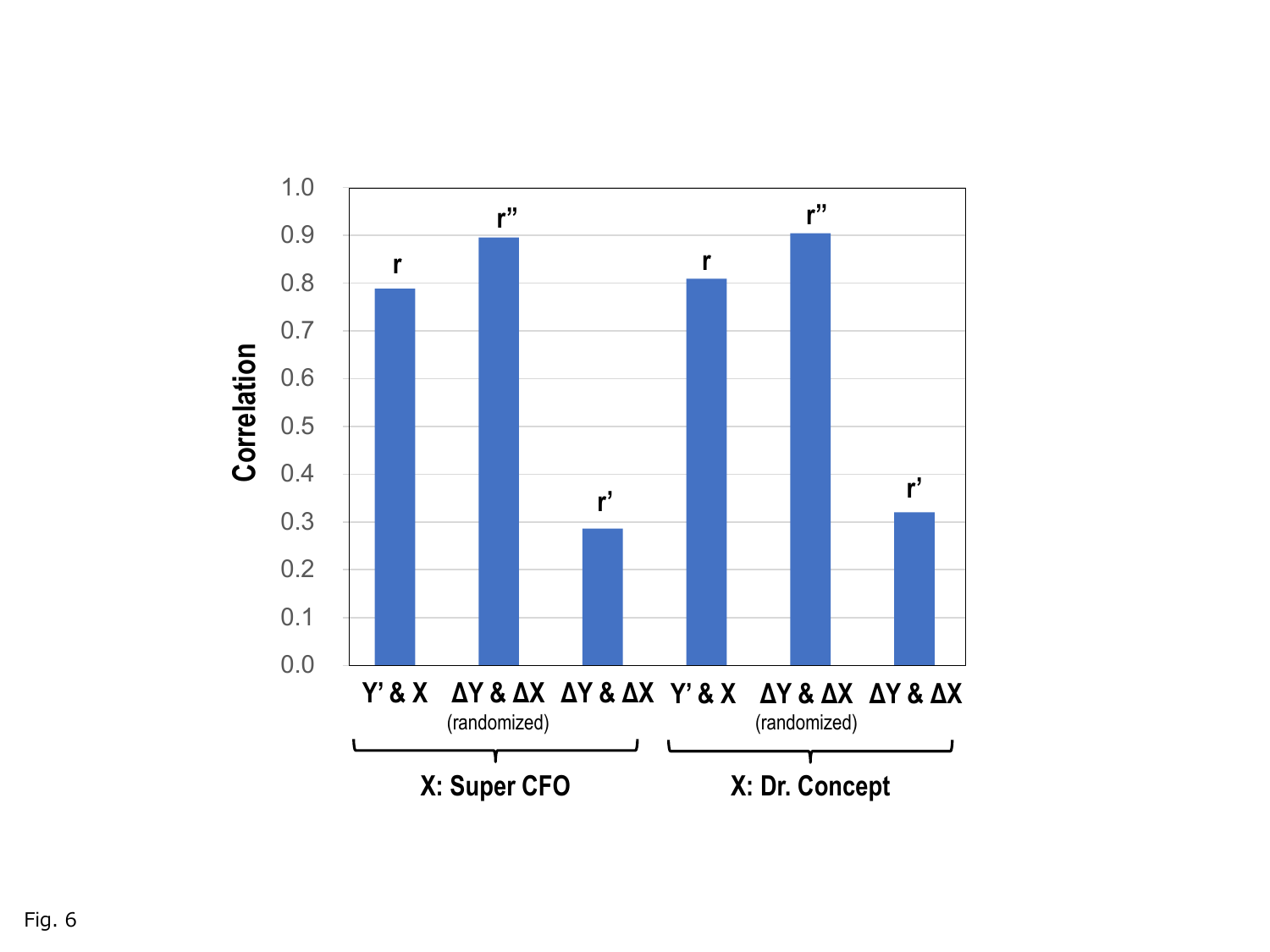}
\caption{Evidence of interference effects in non-commutative semantic composition. Correlation analysis shows that changes in responses induced by sequential operator application cannot be explained by commutative linear composition, revealing interference-like modulation where semantic components are selectively amplified or suppressed.}
\label{fig:6}
\end{figure}

\begin{table*}[t]
  \centering
  \caption{Correlation analysis for detecting interference effects in operator composition.}
  \label{tab:aqi_table3_interference}
  \setlength{\tabcolsep}{4pt}
  
  \vspace{0.5em}
  \begin{adjustbox}{max width=\textwidth}
  \begin{tabular}{@{}ll*{5}{c}@{}}
    \toprule
    \makecell{X\\(Y=Super CFO)}
      & \makecell{Inner Product\\(correlation)}
      & \makecell{1\\Risk \&\\Strategic\\Foresight}
      & \makecell{2\\Investor\\Relations}
      & \makecell{3\\Inter-Organizational\\Collaboration}
      & \makecell{4\\Corporate Finance\\Optimization}
      & \makecell{5\\Investment\\Decision-Making} \\
    \midrule
    \multirow{5}{*}{\makecell{X=\\Super CFO}} & $\mathrm{corr}(Y, X)$ & 0.87 & 0.88 & 0.85 & 0.95 & 0.84 \\
     & $\mathrm{corr}(Y', X)$ & 0.84 & 0.87 & 0.83 & 0.81 & 0.85 \\
     & $\mathrm{corr}(Y, Y')$ & 0.84 & 0.95 & 0.87 & 0.85 & 0.90 \\
     & $\mathrm{corr}(\Delta Y, \Delta X)$ & 0.48 & 0.28 & 0.42 & 0.06 & 0.44 \\
     & $\mathrm{corr}(\Delta Y, \Delta X)_{\mathrm{random}}$ & 0.92 & 0.93 & 0.92 & 0.91 & 0.92 \\
    \addlinespace
    \multirow{5}{*}{\makecell{X=\\Dr.\ Concept}} & $\mathrm{corr}(Y, X)$ & 0.90 & 0.87 & 0.89 & 0.87 & 0.82 \\
     & $\mathrm{corr}(Y', X)$ & 0.82 & 0.79 & 0.81 & 0.85 & 0.85 \\
     & $\mathrm{corr}(Y, Y')$ & 0.85 & 0.90 & 0.87 & 0.88 & 0.92 \\
     & $\mathrm{corr}(\Delta Y, \Delta X)$ & 0.27 & 0.09 & 0.22 & 0.41 & 0.45 \\
     & $\mathrm{corr}(\Delta Y, \Delta X)_{\mathrm{random}}$ & 0.91 & 0.90 & 0.91 & 0.93 & 0.92 \\
    \bottomrule
  \end{tabular}
  \end{adjustbox}

  \vspace{1em}
  
  \begin{adjustbox}{max width=\textwidth}
  \begin{tabular}{@{}ll*{7}{c}@{}}
    \toprule
    \makecell{X\\(Y=Super CFO)}
      & \makecell{Inner Product\\(correlation)}
      & \makecell{6\\Sales \& Client\\Engagement}
      & \makecell{7\\Organizational\\Transformation}
      & \makecell{8\\Human Resource\\Strategy}
      & \makecell{9\\Technology Foresight\\and R\&D Direction}
      & \makecell{10\\Leadership \&\\Mentorship}
      & Average
      & \makecell{Standard\\Deviation} \\
    \midrule
    \multirow{5}{*}{\makecell{X=\\Super CFO}} & $\mathrm{corr}(Y, X)$ & 0.85 & 0.89 & 0.76 & 0.70 & 0.88 & 0.85 & 0.07 \\
     & $\mathrm{corr}(Y', X)$ & 0.77 & 0.81 & 0.62 & 0.69 & 0.79 & 0.79 & 0.07 \\
     & $\mathrm{corr}(Y, Y')$ & 0.88 & 0.87 & 0.72 & 0.84 & 0.86 & 0.86 & 0.05 \\
     & $\mathrm{corr}(\Delta Y, \Delta X)$ & 0.17 & 0.22 & 0.27 & 0.36 & 0.17 & 0.29 & 0.13 \\
     & $\mathrm{corr}(\Delta Y, \Delta X)_{\mathrm{random}}$ & 0.89 & 0.91 & 0.81 & 0.85 & 0.89 & 0.90 & 0.04 \\
    \addlinespace
    \multirow{5}{*}{\makecell{X=\\Dr.\ Concept}} & $\mathrm{corr}(Y, X)$ & 0.83 & 0.80 & 0.76 & 0.82 & 0.89 & 0.85 & 0.04 \\
     & $\mathrm{corr}(Y', X)$ & 0.84 & 0.80 & 0.73 & 0.76 & 0.84 & 0.81 & 0.04 \\
     & $\mathrm{corr}(Y, Y')$ & 0.84 & 0.90 & 0.81 & 0.88 & 0.88 & 0.87 & 0.03 \\
     & $\mathrm{corr}(\Delta Y, \Delta X)$ & 0.52 & 0.36 & 0.38 & 0.20 & 0.29 & 0.32 & 0.12 \\
     & $\mathrm{corr}(\Delta Y, \Delta X)_{\mathrm{random}}$ & 0.92 & 0.90 & 0.87 & 0.88 & 0.92 & 0.90 & 0.02 \\
    \bottomrule
  \end{tabular}
  \end{adjustbox}
\end{table*}

Within the AQI framework, the action of multiple noncommutative operators causes semantic states to evolve not along a single averaged trajectory, but in a manner where different components are selectively reinforced or attenuated. Crucially, this effect does not arise from randomness, but emerges as a necessary consequence of the algebraic structure induced by noncommutative operators. In other words, the observed interference effects demonstrate that expansion of exploration width is accompanied by qualitative diversification in the internal structure of semantic representations, produced through constructive and destructive interactions.

Taken together, the experiments in this chapter show that AQI not only delivers consistent improvements on creative reasoning benchmarks, but also that: (i) even at temperature = 0, swapping the order of operators yields clearly separated output; and (ii) quantum-interference-like structures are observed that cannot be explained by commutative vector composition. These findings support the conclusion that AQI's effectiveness is not a byproduct of increased randomness, but is instead aligned with its core design principle: algebraically extending deterministic dynamics to guarantee a lower bound on exploration.

\FloatBarrier
\section{Discussion}

This study sought to decouple creativity from notions of accidental deviation in meaning generation or exceptional talent, and to reformulate it instead as a dynamical property grounded in noncommutative algebraic structures. The proposed Algebraic Quantum Intelligence (AQI) addresses the tendency of existing large language models (LLMs) to fall into deterministic convergence by introducing order-noncommutative operators that theoretically guarantee a minimum exploration width in semantic space and internalize context-dependent and interference-like dynamics within the generative process.

The primary novelty of this framework lies not in a physical metaphor, but in the fact that abstract structures---noncommutativity and uncertainty principles---are used to make the reproducibility of creative reasoning and the maintenance of diversity designable. Classical theories of creativity, such as Csikszentmihalyi's systems theory or Nonaka's SECI model, have emphasized that creativity does not reside solely within individuals but emerges from dynamic interactions across diverse social and contextual ``fields.'' However, while these theories highlight the importance of interaction, they do not reach the level of quantitatively or dynamically controlling creativity through interaction order or noncommutativity. This study contributes precisely by crossing that boundary into a dynamical description.

Moreover, the noncommutative framework proposed here is consistent with phenomenological accounts of historical bursts of creativity. In contexts such as Renaissance Florence or the publishing culture of Edo-period Japan, dense interactions among individuals with heterogeneous backgrounds, combined with contingencies of order and combination, produced strong effects of semantic interference and reorganization. Such collective creative processes cannot be adequately described as simple sums of individuals or additive networks; rather, they invite reinterpretation as dynamics inherently involving order noncommutativity and interference.

That said, adopting the AQI framework does not automatically guarantee the benchmark results reported in this paper. The evaluated AQI system incorporated substantial design choices and tuning on top of the framework itself. In particular, two aspects were critical. First was the design of individual operators and the overall portfolio of more than 600 operators embodying distinct forms of expertise and perspectives; different operator definitions naturally lead to different outcomes. Second was the design of the S-Generator and H-Generator mechanisms and the associated Hamiltonian control strategy, which dynamically govern the activation and deactivation of operators in response to context. While the specific operator sets and Hamiltonian control strategies implemented for business-oriented applications cannot be disclosed due to intellectual property constraints, the core theoretical structures---noncommutative algebra, the C-value, and the general Hamiltonian form---have been fully described in this paper.

It is also worth noting that this research is informed not only by theoretical considerations but by sustained deployment experience in applied, real-world settings. Although the primary focus of this paper is the foundational algebraic framework, feedback from practical use cases has shaped design priorities and evaluation criteria.

The algebraic approach underlying AQI has clear limitations. Social creativity is deeply entangled with factors such as power structures, cultural frictions, embodied unpredictability, and affective dynamics of place---many of which are inherently unstructured and tacit. Within the current framework, such elements can only be treated implicitly as noise or external conditions, precluding a fully unified theory. Indeed, historical and social bursts of creativity often hinge on serendipitous encounters, spontaneous symbol formation, and the value of unintended deviations.

Precisely because of these limitations, however, the operability of creative dynamics and the theoretical guarantee of a lower bound on exploration offered by AQI have the potential to substantially reshape both creativity research and practice. Applications such as creative AI agents for discussion support, organizational innovation design, and facilitation of emergent brainstorming in social settings suggest a future in which reproducible yet nontrivial idea generation becomes practically actionable. What intentional and structural control of divergence---unattainable through sampling or temperature parameters alone---will yield in real-world deployments remains a key subject for future investigation.

Finally, bridging the tension between theory and real-world practice constitutes an important frontier. Creativity is a domain in which theoretical guarantees and unpredictable reality are intrinsically intertwined, and AQI deliberately leaves this dynamic boundary open. To what extent mathematical models can intertwine with situated knowledge and social practice---extending beyond their current limits---remains an open question. This study aims to serve as a starting point for building such a foundation.

Future work is expected to advance along three directions:
(1) hybrid modeling that integrates complex social and embodied creativity;
(2) iterative validation in real-world settings and dynamic model adaptation; and
(3) the exploration of new epistemic frontiers co-created by creative AI agents and humans.

\section{Conclusion}

This paper proposed Algebraic Quantum Intelligence (AQI) as a new theoretical and computational framework for reproducible machine creativity. The starting point of this work was the observation that, despite their high performance, contemporary large language models tend to fall into effectively deterministic semantic generation dynamics, leading to structural constraints on exploration.
AQI addresses this limitation by introducing noncommutative-operator-based dynamics for semantic generation. Through order effects, interference structures, and uncertainty-type lower-bound constraints, AQI provides a design principle for systematically expanding exploration in semantic space. Crucially, the resulting improvements in creativity do not arise from stochastic noise or heuristic diversification, but from controllable dynamics encoded in the operator structure itself.

Empirically, across ten creative reasoning benchmarks, AQI outperformed fourteen strong baseline models---including GPT-5 and Gemini 3---by an average of 27 T-score points on the Co-Creativity Index (CCI). In addition, operator order-swapping experiments and interference analyses demonstrated that AQI's effects are not attributable to randomness, but instead reflect dynamical properties originating from noncommutative generators.

The contributions of this study can be summarized as follows. First, it formulates the limits of creativity not as a matter of insufficient scale, but as a structural property of semantic generation dynamics. Second, it presents a designable framework based on noncommutative algebra that guarantees a lower bound on exploration. Third, it demonstrates that this framework is implementable on existing computational infrastructure and yields reproducible performance gains.
AQI reframes creativity not as an elusive phenomenon, but as an object of algebraic dynamics that can be analyzed and designed. Going beyond incremental algorithmic improvements, it provides guiding principles for the design of creative reasoning in future AI systems. By freeing semantic generation from a uniquely determined future while preventing collapse into disorder, AQI offers a new foundation for the study of machine creativity.

\section*{Contribution}

Conceptualization: KY2, Methodology: JL, KY2, Formal analysis: JL, KY2, Software: TI, HK, JL, AK, KM, MO, NS, ST1, AT, KY2, KY3, Investigation: TI, JL, HK, KY2, Validation: MH, T, HK, AK, JL, YM1, KM1, KM2, HM, YM2, KM3, YO, TO, NS, MS2, KS, ST2, TT, KY2, Data curation: JL, TI, HK, KY2, Visualization: JL, KY2, Writing -- original draft: HK, JL, KY2, Writing -- review \& editing: All authors, Supervision: KY2, KA, Project administration: YI, KA, AK, KM2, MS1, NS, ST2, AO, ST, KY1, KY2, Resources: NS, MO, KS, Funding acquisition: AO, KY2

\vspace{0.5em}
{\small YM1: Yuki Matsuda, YM2: Yumi Miyazaki, KM1: Kazunori Matsumoto, KM2: Kenichi Matsumura, KM3: Kotaro Murai, MS1: Mirei Saito, MS2: Marie Seki, ST1: Sho Takematsu, ST2: Shun Tanoue, KY1: Kazunori Yanagi, KY2: Kazuo Yano, KY3: Keiko Yano}

\section*{Competing Interests}

The authors are employees of Happiness Planet, Ltd. and have contributed to the development and commercialization of systems related to this work. The reported theoretical formulation and evaluation were conducted independently of commercial considerations.

\end{document}